\documentclass[10pt,twocolumn,letterpaper]{article}

\usepackage[pagenumbers]{cvpr} 

\usepackage{multirow}

\usepackage{xcolor}         
\usepackage{amsmath}
\usepackage{bm}
\usepackage{tabularx}
\usepackage{graphicx}
\usepackage{colortbl}
\usepackage{wrapfig}
\usepackage{pifont}
\usepackage{subcaption}

\usepackage{makecell}    
\usepackage{listings}    
\usepackage{booktabs}    
\usepackage{caption}     
\usepackage{enumitem}    

\newcommand{\cmark}{\ding{51}}%
\newcommand{\xmark}{\ding{55}}%

\definecolor{citecolor}{HTML}{0071bc}
\definecolor{ourscolor}{HTML}{c2d1e5}

\newcommand{\modelname}{UniCAD-MLLM}

\definecolor{cvprblue}{rgb}{0.21,0.49,0.74}
\usepackage[pagebackref,breaklinks,colorlinks,allcolors=cvprblue]{hyperref}

\def\paperID{*****} 
\def\confName{CVPR}
\def\confYear{2026}

\title{UniCAD: A Unified Benchmark and Universal Model for Multi-Modal Multi-Task CAD}
\author{
\centerline{
Jingyuan Chen \thanks{Equal contribution.} \quad
Sheng Jin \footnotemark[1] \quad
Haopeng Sun \quad
Wentao Liu \quad
Chen Qian \quad
}\\
\centerline{
SenseTime Research and Tetras.AI} \\
\centerline{\{chenjingyuan, jinsheng, sunhaopeng, liuwentao, qianchen\}@tetras.ai}
}

\begin{document}

\maketitle

\begin{figure*}
  \centering
  \includegraphics[width=0.9\textwidth]{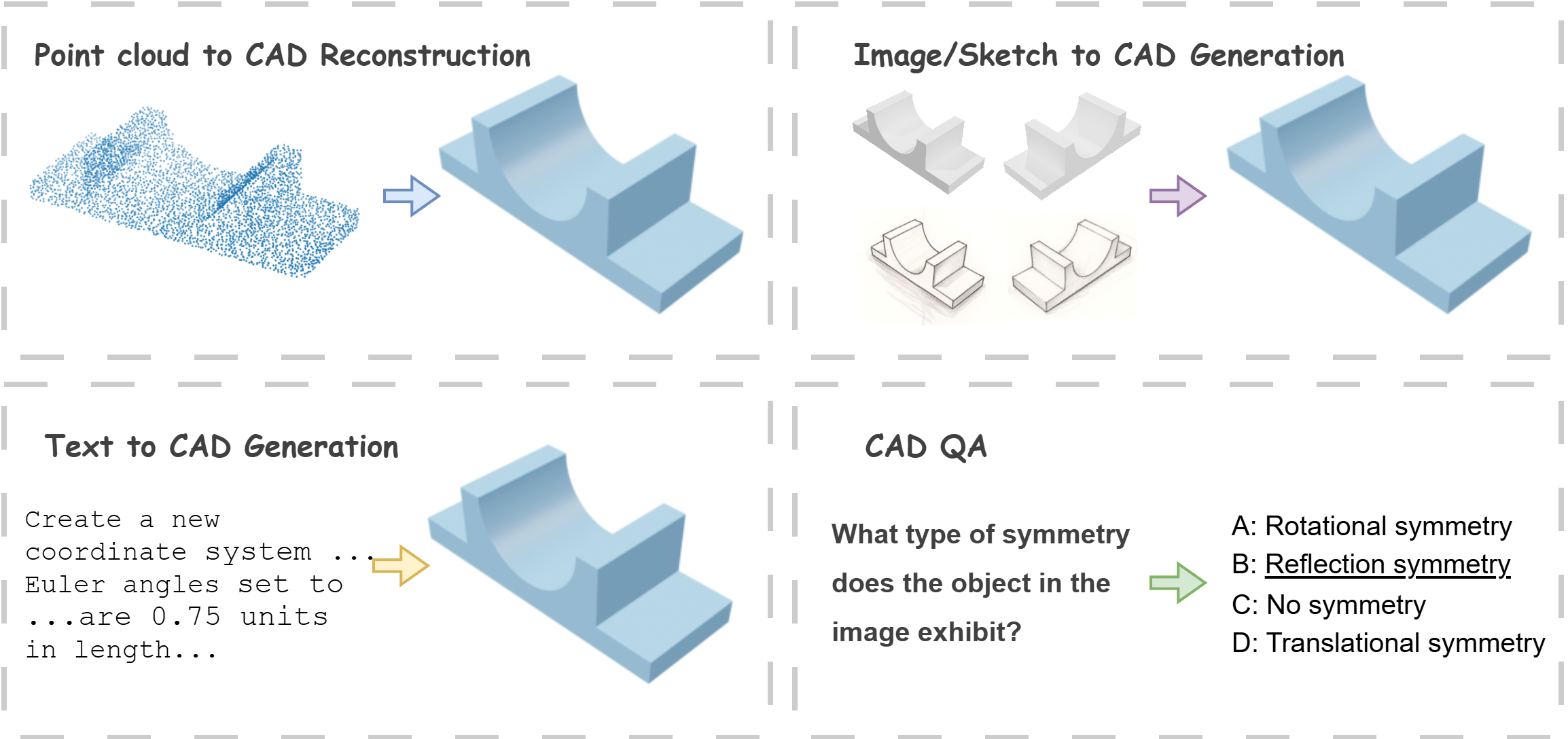}
  \caption{UniCAD unifies (a) point-cloud to CAD reconstruction, (b) image/sketch to CAD generation, (c) text to CAD generation and (d) CAD QA, enabling shared geometric–semantic reasoning across heterogeneous inputs and tasks.}
  \label{fig:motivation}
\end{figure*}

\begin{abstract}
Computer-Aided Design (CAD) underpins modern engineering and manufacturing by enabling the creation of precise, editable 3D models. However, CAD research typically studies tasks in isolation, and multi-modal, multi-task learning for CAD is hindered by the absence of a unified benchmark. To address this gap, we introduce UniCAD, a comprehensive benchmark for multi-modal CAD learning that covers point-to-CAD reconstruction, text/image-to-CAD generation, and CAD question answering across diverse input modalities. Alongside the benchmark, we present UniCAD-MLLM, a universal multi-modal large language model that ingests text, images, sketches, and point clouds and performs these heterogeneous tasks in an end-to-end fashion within a single framework. Extensive experiments on the UniCAD and Fusion360 benchmarks demonstrate that UniCAD-MLLM achieves state-of-the-art performance across all tasks, outperforming existing task-specific and multi-task baselines. We will release the dataset, code, and pretrained models to accelerate future research.
\end{abstract}

\section{Introduction}
Computer-Aided Design (CAD) enables the creation, modification, analysis, and optimization of precise, parametric 3D models and is central to modern engineering and manufacturing workflows~\cite{sarcar2008computer}. CAD representations support iterative design, downstream fabrication, and collaborative workflows, but building CAD models still requires domain expertise and significant manual effort. Recent work aims to lower this barrier by automating CAD generation and manipulation from alternative inputs (\eg, text, images or sketches). Yet, most methods target a single input modality or a single downstream task (reconstruction, generation, or understanding), which limits cross-task knowledge transfer and practical usefulness in real-world design-assistance scenarios where users naturally provide mixed inputs.

A major practical obstacle to unified CAD learning is the lack of a shared, large-scale benchmark that brings together multiple input modalities and diverse tasks under a single, standardized evaluation protocol. Existing datasets and evaluation pipelines typically focus on either a single modality (point clouds, multi-view images, or text) or a single downstream task (\eg, reconstruction, generation, or understanding), making it difficult to compare methods fairly or to develop models that robustly generalize across tasks. 

To address this gap, we introduce UniCAD, a unified benchmark for multi-modal, multi-task CAD learning. UniCAD consolidates and standardizes paired CAD data across text, images, sketches, and point clouds, and defines evaluation protocols and task-specific metrics for several practical tasks: text→CAD, image/sketch→CAD, point-cloud→CAD and CAD QA. By exposing models to heterogeneous modalities and tasks within a single framework, UniCAD encourages the development of systems that learn shared geometric and semantic representations, improves data efficiency via cross-task supervision, and better reflects realistic design scenarios in which users combine sketches, descriptions, and scans.

Complementing the benchmark, we propose UniCAD-MLLM, a multimodal large language model that unifies these tasks. Designing a universal CAD model poses several technical challenges. The input modalities encode complementary but different information: sparse spatial samples in point clouds, dense visual cues in images, and high-level intent in language. Fusing these signals while retaining task-specific detail requires careful encoder design and a robust shared latent. The outputs are also heterogeneous. CAD generation demands precise, executable CAD representations, while CAD understanding (\ie question answering) requires structured retrieval and reasoning over parametric model elements.
UniCAD-MLLM uses modality-specific encoders to ingest text, images/sketches, and point clouds, projects them into a shared geometric–semantic latent, and tackles CAD generation and understanding in a unified manner. A key design decision in our pipeline is to produce CAD outputs as executable, editable Python scripts (rather than commonly used command sequences or only final B-Rep meshes). This representation has several practical advantages: (1) it is directly interpretable and editable by humans, (2) it can be executed and validated programmatically to check for syntactic and semantic correctness, and (3) it integrates naturally with popular CAD toolchains (\eg, CadQuery~\cite{cadquery} workflows), making downstream refinement and deployment more straightforward.

We evaluate UniCAD-MLLM extensively on the UniCAD benchmark against a suite of task-specific and multi-task baselines. Results demonstrate that a unified, multi-task model trained on diverse modalities can outperform specialized systems across generation and understanding. %

In summary, our contributions are threefold:
\begin{itemize}
\item We introduce UniCAD, the first large-scale unified benchmark for multi-modal, multi-task CAD learning, providing standardized data, splits, and evaluation protocols for text$\rightarrow$CAD, image/sketch$\rightarrow$CAD, point-cloud$\rightarrow$CAD, and CAD question answering.
\item We present UniCAD-MLLM, a single end-to-end model that jointly processes text, images/sketches, and point clouds, and unifies CAD generation and understanding within a shared geometric–semantic latent space.
\item We demonstrate that joint multi-modal, multi-task learning consistently outperforms specialized per-task baselines, validating the effectiveness and scalability of unified CAD modeling.
\end{itemize}

\section{Related Work}

\subsection{Representations for CAD}
Existing methods for CAD generation can be broadly categorized based on model representations: boundary representation (B-rep), command sequence, and CAD program representations.

\subsubsection{Boundary Representation (B-rep)}
A B-rep model describes a solid object using a graph structure that encodes both geometric primitives (\eg, parametric curves and surfaces) and topological entities (\eg, vertices, edges, and faces) that together define the model’s boundary~\cite{ansaldi1985brep}. Prior work has explored B-rep generation through different strategies. Early methods relied on predefined template curves or surfaces~\cite{smirnov2021learning, wang2022neural, wang2020pie-net, li2019supervised, sharma2020parsenet}, while PolyGen~\cite{nash2020polygen} represented a special case of B-rep using planar \textit{n-gon} meshes generated via pointer networks and Transformers~\cite{vaswani2017attention}. More recent methods, such as BrepGen~\cite{xu2024brepgen}, directly synthesize complete B-rep models, employing diffusion-based denoising across faces, edges, and vertices.
Although B-rep captures geometric boundaries with high fidelity and serves as the standard format for downstream CAD evaluation, ensuring topological validity (\eg, avoiding self-intersections, gaps, or overlaps) introduces significant modeling complexity. Moreover, B-rep structures are less intuitive for editing and constraint-based reasoning, limiting their applicability in interactive or multi-modal pipelines.

\subsubsection{Command Sequence Representation}
The availability of large-scale parametric CAD datasets~\cite{willis2021fusion360, wu2021deepcad} has catalyzed research on modeling construction command sequences, which capture the procedural history of how a CAD model is built. Each command (\eg, sketch, extrude) encodes parametric and geometric constraints, making this representation both compact and interpretable. Learning-based approaches leverage command histories~\cite{willis2021fusion360, wu2021deepcad, xu2022skexgen, xu2023hnc-cad} and sketch constraints~\cite{seff2020sketchgraphs} to generate editable parametric solids that can be replayed through a modeling kernel to reconstruct the CAD file.  While command-sequence approaches are compelling for their compactness, they are often less expressive for higher-level control structures and may be harder to integrate directly with existing CAD scripting ecosystems.

\subsubsection{CAD Program Representation}
An alternative line of work represents CAD models as executable programs, typically written in Python-based frameworks such as CadQuery~\cite{cadquery}. This representation offers several distinct advantages: it is interpretable, editable, and executable, enabling automatic validation, constraint checking, and integration with human-in-the-loop design workflows. Recent studies~\cite{rukhovich2024cad-recode,xie2025text} have shown that mapping CAD sequences to structured Python scripts can improve both fidelity and generalization, while aligning naturally with modern code-oriented LLMs. In this paradigm, CAD generation becomes a form of code generation, where large vision–language models (VLMs)~\cite{wang2024qwen2-vl, yang2024qwen2} can provide powerful cross-modal reasoning. Our approach follows this direction by generating CAD models as executable Python scripts that unify multiple input modalities within a single framework. Rather than claiming programmatic representations universally outperform other forms, we argue that they are particularly well-suited for multi-task and multi-modal CAD pipelines, where interpretability, editability, and validation are critical for research and industrial applications.


\subsection{CAD Tasks and Benchmarks}

Learning-based CAD research encompasses several key tasks, conditional CAD generation, and CAD question answering (QA), that together aim to endow models with multimodal understanding and controllable design capabilities. However, progress in these tasks has been constrained by fragmented datasets and inconsistent representations. Existing CAD benchmarks (Table~\ref{tab:benchmark}) differ in their representations (B-rep, command sequence, or programmatic code), conditioning modalities (text, image, sketch, or point cloud), and supported tasks, making fair comparison difficult and hindering unified model development. In contrast, UniCAD provides a unified, large-scale, and multimodal benchmark that supports CAD generation and QA tasks jointly, paving the way toward general-purpose CAD understanding and reasoning models.

\begin{table*}[t]
\centering
\caption{Overview of 3D CAD Datasets by CAD Representation, Model Size, Input Modality and Supported Task. * 3D shapes may include B-rep, Mesh or point clouds.
Our proposed UniCAD dataset is the only dataset available that simultaneously supports multi-view images, text, and point cloud conditioned data for CAD reconstruction, generation and understanding.
}
\renewcommand\arraystretch{1.0}
\resizebox{2\columnwidth}{!}
{
\begin{tabular}{cccccccccccc}
\toprule
\multirow{2}{*}{Dataset} & \multirow{2}{*}{Year}  & \multirow{2}{*}{Representation}  & \multirow{2}{*}{Dataset Size}  & \multicolumn{4}{c}{Input Modality} & \multicolumn{3}{c}{Task} \\
\cmidrule(r){5-8}
\cmidrule(r){9-11}
&&& & 3D Shapes$^*$ & Text & Image & Sketch & Recon. & Gen. & Und. \\
\midrule
ABC~\citep{koch2019abc} & 2019 & B-rep & $\sim$1,000,000 & \cmark & \xmark & \xmark & \xmark & \cmark & \xmark & \xmark \\ 
CC3D-Ops~\citep{dupont2022cadops} & 2022 & B-rep & $\sim$37{,}000 & \cmark & \xmark & \xmark & \xmark & \cmark & \xmark & \xmark \\ 
DeepCAD~\citep{wu2021deepcad} & 2021 & Command Sequence & 179{,}133 & \cmark & \xmark & \xmark & \xmark & \cmark & \xmark & \xmark \\ 
Fusion360~\citep{willis2021fusion360} & 2021  & Command Sequence & 8{,}625 & \cmark & \xmark & \xmark & \xmark & \cmark & \xmark & \xmark \\ 
CADParser~\citep{zhou2023cadparser} & 2023 & Command Sequence & $\sim$40{,}000 & \cmark & \xmark & \xmark & \xmark & \cmark & \xmark & \xmark \\ 
CAD-Recode~\citep{rukhovich2024cad-recode} & 2024 & CAD Programs  & $\sim$1,000,000 & \cmark & \xmark & \xmark & \xmark & \cmark & \xmark & \xmark \\ 
\midrule
Text2CAD~\citep{khan2024text2cad} & 2024 & Command Sequence & $\sim$158{,}000 & \xmark & \cmark & \xmark & \xmark & \xmark & \cmark & \xmark \\
Query2CAD~\citep{badagabettu2024query2cad} & 2024 & CAD Programs  & 57 & \xmark & \cmark & \xmark & \xmark & \xmark & \cmark & \xmark \\
Text2CADQuery~\citep{xie2025text} & 2025 & CAD Programs & $\sim$170{,}000 & \xmark & \cmark & \xmark & \xmark & \xmark & \cmark & \xmark \\
\midrule
Img2CAD~\citep{you2024img2cad} & 2024 & Command Sequence & 4{,}574  & \xmark & \xmark & \cmark & \xmark & \xmark & \cmark & \xmark \\
ABC-mono~\citep{chen2024img2cad}
  & 2024 & Command Sequence & 208{,}853 & \xmark & \xmark & \cmark & \cmark & \xmark & \cmark & \xmark \\
OpenECAD~\citep{yuan2024openecad} & 2024 & Command Sequence & $\sim$200{,}000 & \xmark & \xmark & \cmark & \xmark & \xmark & \cmark & \xmark \\
Free2CAD~\citep{li2022free2cad} & 2022 & Command Sequence & $\sim$210{,}000 & \xmark & \xmark & \xmark & \cmark & \xmark & \cmark & \xmark \\
\midrule
SGP-Bench~\citep{qiu2024can} & 2025 & Command Sequence & 2,400  & \xmark & \xmark & \cmark & \xmark & \xmark & \xmark & \cmark \\
\midrule
OmniCAD~\citep{xu2024cad} & 2024 & Command Sequence & 453{,}220 &  \cmark &  \cmark & \cmark &  \xmark &  \cmark & \cmark & \xmark \\
LLM4CAD~\citep{sun2025large} & 2025 & CAD Programs  & $\sim$5000 & \xmark & \cmark & \cmark & \cmark & \xmark & \cmark & \xmark \\ 
\rowcolor{ourscolor}
UniCAD (Ours) & 2025 & CAD Programs  & 1,448,150 & \cmark & \cmark & \cmark & \cmark & \cmark & \cmark & \cmark\\
\bottomrule
\end{tabular}
}
\label{tab:benchmark}
\end{table*}

\textbf{CAD Generation}
focuses on synthesizing parametric 3D models that adhere to constraints specified by inputs such as point clouds, images, or text. DeepCAD~\cite{wu2021deepcad} introduced a foundational approach by learning a latent space of CAD command sequences in an unconditional setting, but lacks support for conditional generation. Subsequent methods enable conditioning on individual modalities, including point clouds~\cite{uy2022point2cyl,dupont2024transcad}, images~\cite{you2024img2cad,alam2024gencad,chen2025cadcrafter}, sketches~\cite{li2020sketch2cad,li2022free2cad}, and text~\cite{khan2024text2cad,badagabettu2024query2cad}. Despite continued progress, these approaches are largely modality-specific and rely on disjoint datasets, limiting cross-modal generalization. In contrast, multimodal CAD generation has been comparatively underexplored. Early efforts such as CAD-GPT~\cite{wang2025cad-gpt} combine image and text inputs, but still lag behind specialized single-modality systems in reconstruction fidelity and controllability. A concurrent work, Cadrille~\cite{kolodiazhnyi2025cadrille}, also investigates multimodal CAD reconstruction, further highlighting growing interest in unified multimodal CAD modeling.

\textbf{CAD Question Answering (CAD QA)}
aims to assess a model’s ability to reason about geometric structure and procedural design logic, including tasks such as constraint identification, feature counting and parameter retrieval. While large-scale 3D QA benchmarks exist for meshes and point clouds, QA over parametric CAD representations remains notably underdeveloped. Among the limited efforts, SGP-Bench~\cite{qiu2024can} evaluates a custom CAD domain-specific language for 2D/3D shapes.

\subsection{Large Multimodal Models on CAD}
Recent advances in Large Multimodal Models (LMMs)~\cite{achiam2023gpt,bai2023qwenvl,wang2024qwen2-vl,lu2024deepseekvl} have inspired their application to parametric CAD modeling. Recent works increasingly leverage pretrained LMM architectures for cross-modal reasoning and design synthesis. Img2CAD~\cite{you2024img2cad} employs a vision–language model to infer global discrete structures from images before predicting fine-grained attributes, while Text2CAD~\cite{khan2024text2cad} leverages LMMs for auto-regressive CAD sequence generation. CAD-MLLM~\cite{xu2024cad} further explore multimodal alignment between textual, visual, and geometric modalities for CAD understanding and synthesis. In parallel, LMM-driven agentic systems such as CAD-Assistant~\cite{mallis2024cad-assistant}, CADCodeVerify~\cite{alrashedy2024generating}, and Seek-CAD~\cite{li2025seek} focus on generating or refining executable CAD programs by prompting off-the-shelf LMMs (\eg GPT-4o~\cite{hurst2024gpt-4o} and Gemini~\cite{team2023gemini}). These approaches highlight the potential of LMMs for procedural CAD generation, yet they predominantly rely on generic foundation models without dedicated design for scalable, structured multimodal CAD reasoning. In contrast, \modelname{} advances this frontier by unifying three modalities within a shared framework for controllable CAD generation and understanding.

\section{Creation of UniCAD Dataset}
\label{sec:create_dataset}

Existing CAD datasets predominantly focus on either geometric representations or modeling sequences, rarely providing aligned multi-modal signals or supporting multi-task learning. The ABC dataset~\cite{koch2019abc} offers over one million CAD models but is limited to B-rep geometry and lacks modeling histories. Fusion360 Reconstruction~\cite{willis2021fusion360} provides human-authored construction sequences but is constrained in scale, comprising only 8,625 designs. Other datasets, such as DeepCAD~\cite{wu2021deepcad}, Text2CAD~\cite{khan2024text2cad}, and CAD-Recode~\cite{rukhovich2024cad-recode}, include modeling sequences yet remain single- or limited-modality, restricting their applicability for unified multimodal learning. To bridge this gap, we introduce a large-scale, fully aligned multimodal CAD dataset that integrates text descriptions, CAD programs, rendered images, sketches, point clouds, and QA pairs. This unified design supports diverse research directions, including text-to-CAD, image/sketch-to-CAD, point-cloud-to-CAD, and CAD-based question answering, enabling both multi-modal understanding and generation within a single benchmark.

\begin{figure}[t]
\centering
\includegraphics[width=0.5\textwidth]{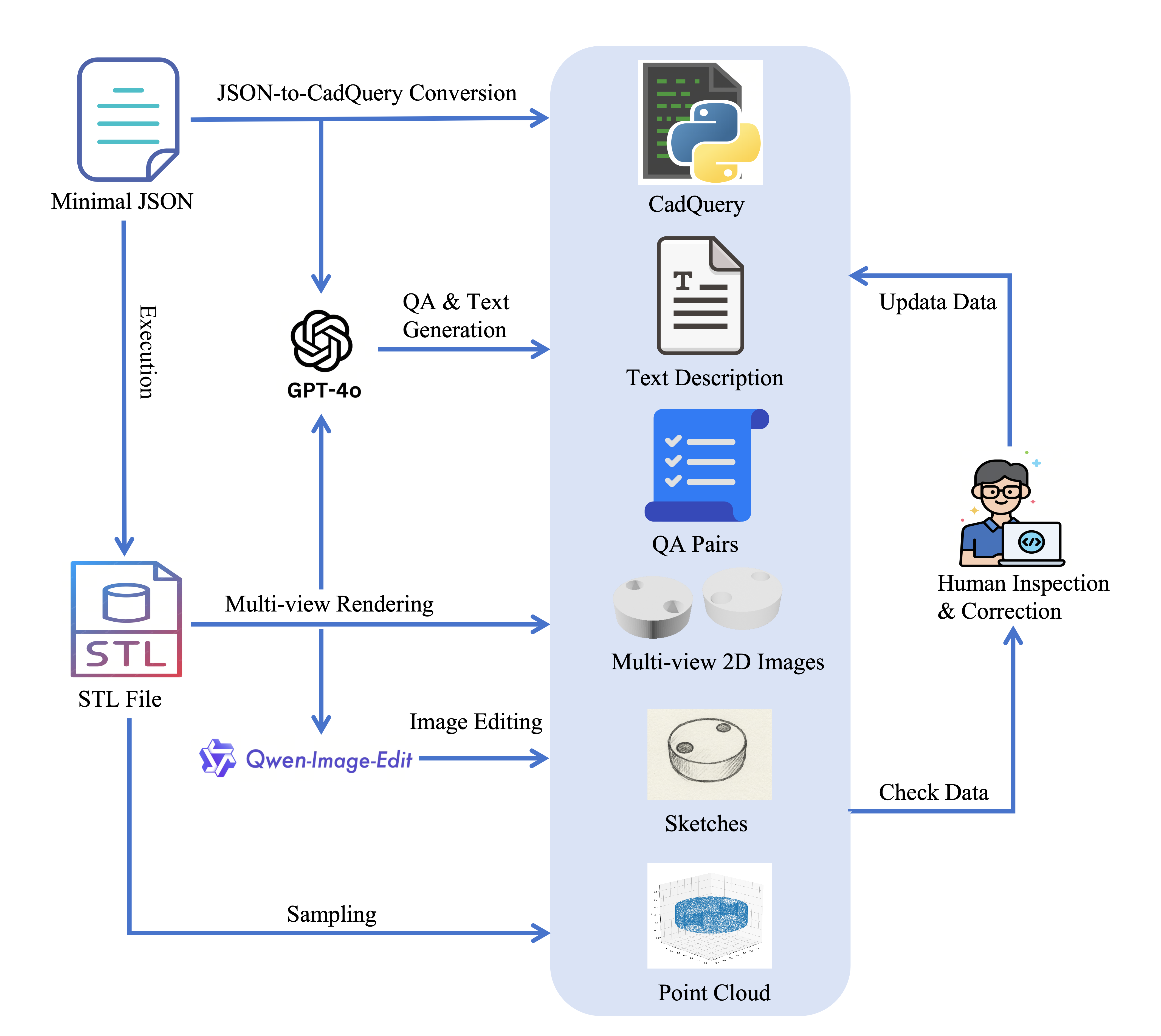}
\caption{Overview of our data generation pipeline.}
\label{fig:data_pipeline}
\end{figure}

\begin{figure*}[t]
\centering
\includegraphics[width=0.9\textwidth]{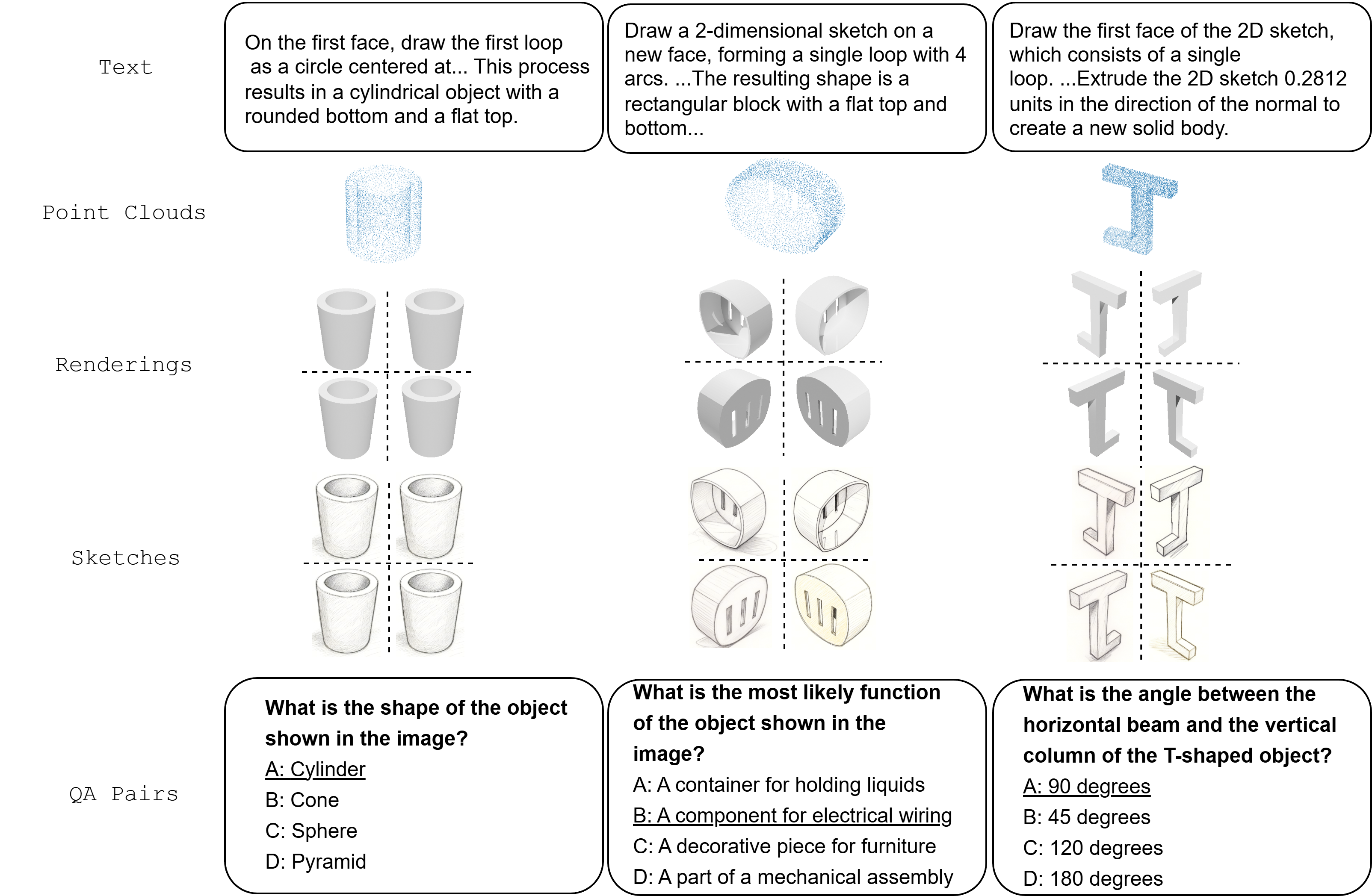}
\caption{Examples from UniCAD. Each sample consists of aligned multi-modal data, including textual descriptions, point clouds, multi-view renderings, sketches, and QA annotations. While the full dataset provides images from eight viewpoints per CAD model, we display 4 randomly selected views for clarity.}
\label{fig:dataset_gallery}
\end{figure*}

Our dataset builds upon three foundational sources: DeepCAD~\cite{wu2021deepcad}, Text2CAD~\cite{khan2024text2cad}, and CAD-Recode~\cite{rukhovich2024cad-recode}. DeepCAD provides 178K parametric CAD models represented as sketch–extrude command sequences. Text2CAD further augments a subset of these models with 600K+ natural language descriptions, including both high-level semantic summaries and step-by-step modeling instructions. CAD-Recode contributes over 1M synthetically generated CAD programs, where sketch and operation parameters are randomly sampled under topological and geometric heuristics to ensure controllable structure and feature diversity. 

We extend these datasets with additional multimodal annotations to improve scale and diversity. To enable fair comparison with prior work, we preserve the original DeepCAD/Text2CAD test split (8,046 models) for CAD generation evaluation. This benchmark is further enriched with multi-view renderings, sketches, and QA pairs to support comprehensive multimodal assessment as shown in Fig.~\ref{fig:dataset_gallery}.

\subsection{JSON-to-CadQuery Conversion}
We convert all CAD designs into CadQuery~\cite{cadquery} scripts, a Python-based parametric CAD framework that is executable, interpretable, and lightweight. Compared to FreeCAD or OpenSCAD, CadQuery provides a clean API (\eg, \texttt{box()}, \texttt{circle()}, \texttt{extrude()}) and can be executed without a full CAD environment, making it ideal for integration with large language models (LLMs) that natively generate Python code. This choice ensures that each sample in our dataset is both verifiable and editable, aligning with the practical needs of design automation and human-in-the-loop CAD workflows. Text-to-CADQuery~\cite{xie2025text} uses large language models (LLMs) to convert structured JSON descriptions of modeling operations into executable CadQuery code. While scalable, this LLM-based method is non-deterministic and error-prone (85\% success rate reported by the authors), often generating invalid sketches or missing operations. In contrast, we developed a deterministic JSON-to-CadQuery translator via a rule-based Python script. Each JSON operation maps directly to a corresponding CadQuery API call, guaranteeing one-to-one correspondence, syntactic validity, and reproducibility. Our converter achieves 100\% execution success, producing geometrically consistent CAD models with no hallucinations. To validate geometric faithfulness, we compare the converted models with DeepCAD ground-truth models: 99.2\% of samples achieve $\ge$98\% voxel overlap.
This rule-based design offers superior reliability and interpretability, making it well-suited for large-scale CAD dataset annotation and multi-modal learning.

\subsection{Point Cloud Generation}
To bridge geometric and perception-based learning, we generate point clouds by uniformly sampling points from each CAD mesh surface. 
Each point is annotated with normal vectors derived from the underlying geometry, providing an additional modality for point-based reconstruction models.

\subsection{Synthetic Rendering}
For each CAD model, we render multi-view synthetic image renderings from eight fixed perspectives using the CadQuery toolchain. Eight rendering views include:  Front, Left, Right, Top, and four isometric views from $(\pm s,\pm s,s)$.
The rendering process employs a neutral background, consistent lighting, and standardized camera poses to ensure cross-model comparability. 
These images serve as the visual grounding for tasks such as image-conditioned CAD generation.

\subsection{Sketch Generation}
Previous works~\cite{you2024img2cad,chen2024img2cad} have often relied on heuristic methods such as edge detection to produce sketch-like renderings, yet these heuristic methods yield unrealistic or noisy results. We instead employ the open-source Qwen-Image-Edit model~\cite{wu2025qwen}, which jointly performs semantic and appearance-level editing. 
Starting from synthetic renderings, we apply controlled image editing to produce  sketches that mimic hand-drawn line art consistent with real design drafts. The corresponding prompts are provided in the supplementary.
This approach significantly improves the diversity and realism of image modalities while maintaining geometric consistency with the source CAD. 

\subsection{CAD Question Answering (CAD QA)}
To construct the CAD QA subset, we automatically generate question–answer pairs grounded in rendered CAD images and symbolic CAD programs, enabling evaluation of joint geometric perception and symbolic–procedural reasoning.
We adopt a semi-automatic annotation pipeline. Given multi-view CAD renderings and their corresponding modeling programs, GPT-4o simultaneously generates questions and answers in a 4-way multiple-choice format. The generated questions target both visual–geometric understanding and program-level reasoning (\eg, “What shape results if the second extrusion is removed?”), with answers grounded in the CAD models. To validate annotation quality, we conduct human verification on a subset of samples. A study on 500 randomly sampled QA pairs shows $>$92\% agreement between model-generated and human-authored answers, confirming the reliability of the automatic annotations. We next conduct human-in-the-loop filtering, whereby annotators inspect and amend generated answers to guarantee correctness.

\subsection{Dataset Statistics}

\begin{figure}[t]
\centering
\includegraphics[width=0.49\textwidth]{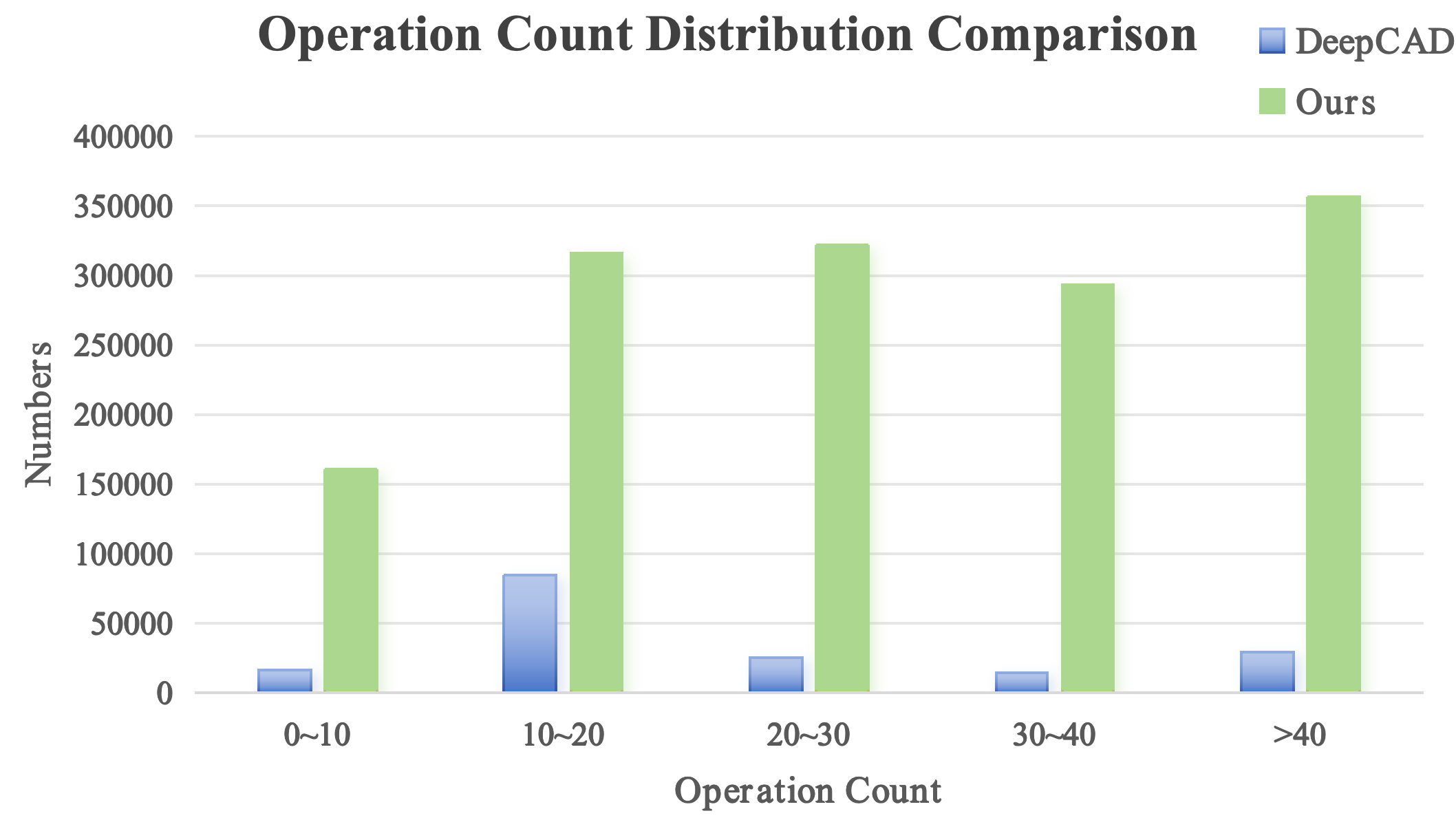}
\caption{Comparison of operation counts per CAD model between UniCAD (Ours) and DeepCAD.}
\label{fig:data_stat}
\end{figure}

UniCAD contains 1,448,150 CAD models, making it one of the largest multi-modal CAD datasets to date. Each sample is paired with a CadQuery program, textual description, multi-view images and sketches, and point clouds, enabling joint learning across diverse 3D representations.
As shown in Fig.~\ref{fig:data_stat}, UniCAD exhibits significantly higher CAD operation counts per model than DeepCAD~\cite{wu2021deepcad}, indicating greater geometric and topological complexity. This increased structural richness presents a more challenging and representative benchmark for real-world CAD understanding and generation tasks.
\begin{figure*}[th]
\centering
\includegraphics[width=0.75\textwidth]{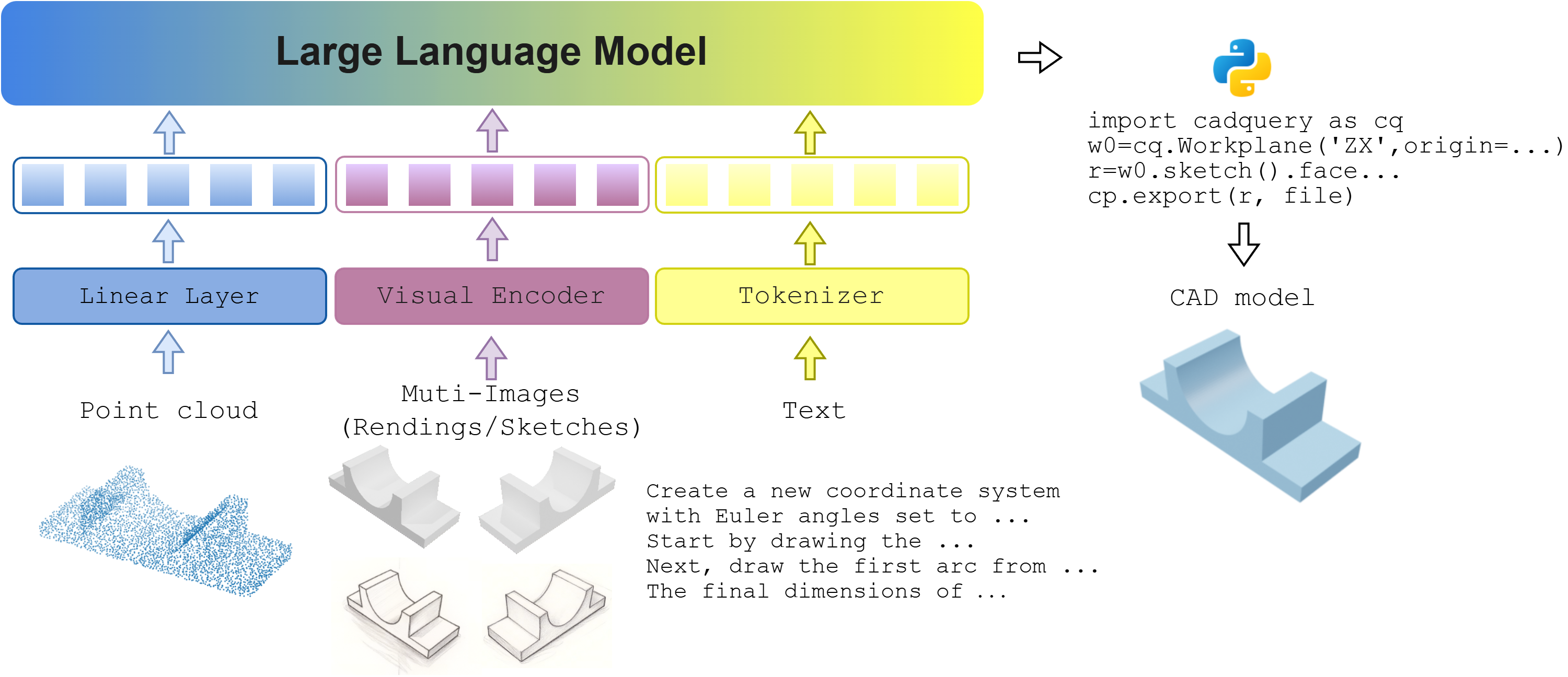}
\caption{Overview of UniCAD-MLLM architecture. The model processes three distinct input modalities, each distinguished by a unique color code: point clouds are handled via a trainable projection layer, images are encoded using the visual encoder, and text is fed into the tokenizer. Its outputs are CADQuery programs for CAD reconstruction and generation tasks, and textual responses for CAD QA tasks.}
\label{fig:pipeline}
\end{figure*}

\section{Method}

\subsection{Problem Formulation}

We formulate CAD modeling as a sequence generation task over a programmatic representation. A CAD model is represented by an executable program $\tau = (t_1, t_2, \dots, t_n)$, where $t_i$ corresponds to a CADQuery code token. Our goal is to learn a single, multimodal model $\pi_\theta$ that can perform diverse CAD-related tasks conditioned on a multimodal query $q$, which can be a combination of text $X_t$, a set of images $X_v$, and a point cloud $X_p$. 

\textbf{(1) CAD Reconstruction and Generation.}
For generation tasks, the model takes a query $q = (X_t, X_v, X_p)$ specifying the desired object and generates a complete, executable CAD program. The model's output is the program $\tau$ that satisfies the geometric and semantic constraints of the query:
$\tau = \pi_\theta(q).$

\textbf{(2) CAD Question Answering.}
For reasoning tasks, the model is given a query $q$ and an existing CAD program $\tau$. It then generates a textual answer $a$ by jointly reasoning over the query and the program's symbolic structure and implicit geometry:
$a = \pi_\theta(q, \tau).$

\subsection{UniCAD-MLLM Architecture}
As illustrated in Fig.~\ref{fig:pipeline}, UniCAD-MLLM extends Qwen2-VL-2B~\cite{wang2024qwen2-vl}, a powerful multimodal large language model (MLLM), to handle CAD tasks. We integrate a lightweight, trainable point cloud projection module and fine-tune the model to decode CAD programs and textual answers. Notably, UniCAD-MLLM does not mandate the presence of all modalities, accommodating real-world data scarcity. Through a unified prompt framework, each modality is treated as an optional token module; missing inputs (text/image/point cloud) are replaced by generic instructions, enabling flexible multimodal reasoning.

\textbf{(1) Multimodal Encoders.}
UniCAD-MLLM processes three distinct input modalities.
For a textual description $X_t$, we use the MLLM's native text tokenizer. For a set of $k$ multi-view images $X_v = \{x_v^i\}_{i=1}^k$, we extract visual features using the pretrained MLLM vision encoder.
For 3D geometry, we introduce a dedicated point cloud encoder. Given a point cloud $X_p \in \mathbb{R}^{N \times 3}$ with $N$ points, our module maps it to a fixed-size set of descriptive embedding vectors $\mathbf{E}_p \in \mathbb{R}^{N_p \times d_{\text{model}}}$, where $N_p \ll N$ and $d_{\text{model}}$ is the model's hidden dimension. This projection module, inspired by CAD-Recode~\cite{rukhovich2024cad-recode}, consists of three stages:
(1) Furthest Point Sampling (FPS) to select a representative subset of $N_p$ points from $X_p$.
(2) Fourier positional encoding~\cite{zhao2023michelangelo} applied to the 3D coordinates of the sampled points to capture high-frequency details.
(3) A linear projection layer that maps the resulting features into the $d_{\text{model}}$-dimensional embedding space. This module is lightweight and trained end-to-end with the rest of the model.

\textbf{(2) Unified Input Representation.}
Given a multimodal query $q = (X_t, X_v, X_p)$, we first encode each modality into a sequence of embedding vectors: $\mathbf{E}_t$ for text, $\mathbf{E}_v$ for images, and $\mathbf{E}_p$ for the point cloud. These sequences are then concatenated to form a single, unified prompt sequence $\mathbf{E}_{\text{prompt}} = \text{Concat}(\mathbf{E}_t, \mathbf{E}_v, \mathbf{E}_p)$, where $\text{Concat}(\cdot)$ joins the sequences along the token dimension. This unified representation serves as the input to the MLLM's decoder.

\textbf{(3) Autoregressive Decoding for Generation and QA.}
The core of UniCAD-MLLM is an autoregressive language decoder, which generates outputs token by token, conditioned on the input embeddings.
For CAD generation and reconstruction, the decoder models the probability of the next program token $t_i$ given the preceding tokens and the prompt embeddings $p(t_i | t_{<i}, \mathbf{E}_{\text{prompt}}).$
The final program $\tau$ is generated by sampling from this distribution.
For question answering, the input to the decoder is a concatenation of the prompt embeddings $\mathbf{E}_{\text{prompt}}$ and the token embeddings of the given CAD program $\mathbf{E}_\tau$. The decoder then generates the textual answer $a$ by reasoning over this combined context.
Both tasks are optimized jointly under the standard cross-entropy loss for autoregressive language modeling. By fine-tuning the language model and the new projection module, UniCAD-MLLM enhances its cross-modal reasoning and geometric understanding capabilities.

\subsection{Implementation Details}

\noindent\textbf{Model Configuration.}
UniCAD-MLLM inherits the text and image processing capabilities of Qwen2-VL. For the 3D modality, we sample $N_p = 256$ points from each point cloud via FPS. These points are then embedded into the model's hidden dimension of $d_{\text{model}} = 1536$ using a single linear layer. During training, we randomly sample the number of input image views (ranging from one to four views) for data augmentation, while for evaluation, we use four fixed-pose rendered views.

\noindent\textbf{Training Setup.}
We initialize our model from Qwen2-VL-2B-Instruct, training the new point cloud projection module from scratch while fine-tuning the language model's parameters. The model is trained for 190k steps using the AdamW optimizer with a peak learning rate of $2 \times 10^{-4}$. The training was performed on 8 NVIDIA A800 GPUs, using a per-GPU batch size of 2 with 4 gradient accumulation steps to achieve an effective batch size of 64. The entire process consumed approximately 800 GPU hours.

\begin{table*}[t]
\caption{Results on UniCAD test set. The best results are \textbf{bold}. Our \modelname{} trained jointly on three modalities outperforms all existing modality-specific methods. For fair comparisons, all results are obtained without RL fine-tuning or test-time sampling. $\uparrow$ means higher is better. $\downarrow$ means lower is better.
}
\centering
\resizebox{\linewidth}{!}{%
\setlength{\tabcolsep}{15pt}
\begin{tabular}{l|cc|cc|cc|cc|c}
\toprule
\multirow{2}{*}{Method} &  \multicolumn{2}{c|}{Point Cloud}  & \multicolumn{2}{c|}{Text} & \multicolumn{2}{c|}{Multi-view Images} & \multicolumn{2}{c|}{Multi-view Sketches} & CAD QA\\
 & CD\textdownarrow & IoU\textuparrow  & CD\textdownarrow & IoU\textuparrow & CD\textdownarrow & IoU\textuparrow & CD\textdownarrow & IoU\textuparrow & 
 Acc\textuparrow  \\
\midrule
PointNet\textrightarrow DeepCAD~\cite{wu2021deepcad}  & 9.64 & 46.7  & - & -  & - & - & - & - & - \\
Point-BERT\textrightarrow HNC-CAD~\cite{xu2023hnc-cad}  & 8.64 & 65.3 & - & - & - & - & - & - & - \\
MultiCAD~\cite{ma2023multicad}  & 8.09 & - & - & - & - & - & - & - & - \\
TransCAD~\cite{dupont2024transcad}  & 4.51 & 65.5 & - & - & - & - & - & - & -\\
PrismCAD~\cite{lambourne2022prismcad}  & 4.28 & 72.1 & - & - & - & - & - & - & -\\
Point2Cyl~\cite{uy2022point2cyl}  & 4.27 & 73.8 & - & - & - & - & - & - & -\\
CAD-Diffuser~\cite{ma2024cad-diffuser}  & 3.02 & 74.3 & - & - & - & - & - & - & -\\
CAD-SIGNet~\cite{khan2024cad-signet}  & 0.29 & 77.3 & - & - & - & - & - & - & - \\
CAD-Recode~\cite{rukhovich2024cad-recode}  & 0.18 & 87.1 & - & - & - & - & - & - & - \\
\midrule
BERT\textrightarrow DeepCAD~\cite{wu2021deepcad}  & - & - &  32.8 & -  & - & - & - & - & -\\
Text2CAD~\cite{khan2024text2cad}& - & - &  0.37 & 71.5  & - & - & - & - & -\\
\midrule
DINOv2\textrightarrow HNC-CAD~\cite{xu2023hnc-cad}  & - & - & - & - & 2.08 & - & - & - & -\\
DINOv2\textrightarrow DeepCAD~\cite{wu2021deepcad}  & - & - & - & - & 1.13 & -  & - & - & -\\
CADCrafter~\cite{chen2025cadcrafter}  & - & - & - & - & 0.26 & -  & - & - & -\\
Cadrille~\cite{kolodiazhnyi2025cadrille} & 0.18 & 87.1 & \textbf{0.20} & 82.1  & 0.18 & 86.1  & - & - & -\\
\midrule
GPT-4o~\cite{hurst2024gpt-4o}  & -  &  - & - & -  &  - & -  & - &  - &  0.77 \\
Doubao-Seed-1.6~\cite{ByteDanceSeed2025}   & - & - & - &  - &  - & -  &  - & - & 0.85 \\
\rowcolor{ourscolor}
\textbf{\modelname} & \textbf{0.17} & \textbf{89.7} & \textbf{0.20} & \textbf{86.3} & \textbf{0.17} & \textbf{89.8} & \textbf{0.22} & \textbf{84.2} & \textbf{0.90} \\
\bottomrule
\end{tabular}%
}
\label{tab:unicad-sota}
\end{table*}

\section{Experiments}

\subsection{Evaluation Metrics}
We adopt two standard evaluation metrics: Chamfer Distance (CD) and Volumetric Intersection over Union (IoU). These metrics are used consistently across all methods to evaluate the quality of the generated 3D shapes.

\textbf{Chamfer Distance (CD):} Measures the geometric discrepancy between the generated and ground-truth 3D models~\citep{fan2017point}. For each STL mesh, we normalize its scale, uniformly sample 8,192 surface points, and compute the average squared distance between nearest neighbors across both point clouds:
\begin{equation}
\mathrm{CD}(P, Q) = \frac{1}{|P|} \sum_{p \in P} \min_{q \in Q} \|p - q\|^2 + \frac{1}{|Q|} \sum_{q \in Q} \min_{p \in P} \|q - p\|^2,
\label{eq:chamfer}
\end{equation}

\noindent where $P$ and $Q$ are point sets sampled from the predicted and ground-truth shapes, respectively. Lower values indicate higher geometric fidelity. Values are multiplied by 10\textsuperscript{3}.

\textbf{Volumetric Intersection over Union (IoU):} To assess 3D shape similarity, we compute the volumetric Intersection over Union (IoU) between voxelized versions of the predicted and ground-truth meshes~\citep{wu20153d}. Both meshes are first normalized to fit within the unit cube $[0, 1]^3$, then voxelized using a resolution of $0.02$. Let $V_1$ and $V_2$ denote the occupied voxel grids. IoU is computed as:
\begin{equation}
    \mathrm{IoU} = \frac{|V_1 \cap V_2|}{|V_1 \cup V_2|}
\end{equation}
This metric captures global shape overlap and is particularly useful for assessing coarse structural fidelity. Padding is applied to align voxel grid sizes when necessary.

\subsection{Experimental Results}

\paragraph{Main Results on UniCAD}
Tab.~\ref{tab:unicad-sota} summarizes results on the UniCAD benchmark across four input modalities (point clouds, text, multi-view images, sketches) and the CAD-QA task. Note that the UniCAD test set is identical to that of DeepCAD/Text2CAD, and some state-of-the-art results are sourced from~\cite{kolodiazhnyi2025cadrille}.
Our model achieves superior overall performance through a unified multimodal objective, outperforming prior methods that are typically modality-specialized. Furthermore, our results demonstrate that cross-task and cross-modal supervision provide complementary signals that significantly enhance data efficiency. Notably, \modelname{} achieves the highest geometric reconstruction fidelity with CD of 0.17–0.22 and IoU up to 89.8\% on multi-view images. On the CAD QA subset, \modelname{} reaches 90.0\% accuracy, outperforming general-purpose VLM baselines such as GPT-4o~\cite{hurst2024gpt-4o} and Doubao-Seed-1.6~\cite{ByteDanceSeed2025} by a substantial margin, demonstrating superior joint visual–procedural reasoning.

\begin{table}
    \centering
    \captionof{table}{Results of CAD reconstruction from a \textit{single} image on the UniCAD dataset.
    }
    \resizebox{0.75\linewidth}{!}{
    \setlength{\tabcolsep}{15pt}
    \begin{tabular}{l|cc}
    \toprule
    \multirow{2}{*}{Method} &  \multicolumn{2}{c}{Single-view Image} \\
    & CD\textdownarrow & IoU\textuparrow \\
    \midrule
    GPT-4o~\cite{hurst2024gpt-4o, wang2025cad-gpt} & 62.6 & - \\
    CAD-GPT~\cite{wang2025cad-gpt} & 9.77 & -   \\
    DINOv2\textrightarrow HNC-CAD~\cite{xu2023hnc-cad} & 2.14 & - \\
    Img2CAD~\cite{chen2024img2cad} & 1.60 & - \\
    DINOv2\textrightarrow DeepCAD~\cite{wu2021deepcad} & 1.26 & - \\
    CADCrafter~\cite{chen2025cadcrafter} & 0.72 & -  \\
    Cadrille~\cite{kolodiazhnyi2025cadrille} & 0.21 & 81.7  \\
    \rowcolor{ourscolor}
    \textbf{\modelname} & \textbf{0.17} & \textbf{87.6} \\
    \bottomrule
\end{tabular}}
\label{tab:single-image}
\end{table}

\paragraph{Single-view CAD Reconstruction}
We further evaluate single-image CAD reconstruction to compare with image-only baselines (Tab.~\ref{tab:single-image}). Note that some results are borrowed from~\cite{kolodiazhnyi2025cadrille}. Despite being trained on multimodal inputs rather than single images alone, \modelname{} outperforms all existing methods, improving median CD from 0.21 to 0.17 and IoU from 81.7\% to 87.6\% over Cadrille~\cite{kolodiazhnyi2025cadrille}. As expected, single-view reconstruction remains slightly more ambiguous than multi-view input (87.6\% vs. 89.8\% IoU), but the gap remains small, indicating robust visual reasoning and shape completion ability. 
We additionally compare to prompting-based CAD systems (\eg GPT-4o and CAD-GPT~\cite{wang2025cad-gpt}), which can be prompted to directly generate CAD code from images, but exhibits a high Chamfer Distance (62.6 CD in Tab.~\ref{tab:single-image}), confirming that general-purpose VLMs struggle with the structural and syntactic demands of executable CAD generation.

\paragraph{Zero-shot Generalization Assessment on Fusion360}
To assess cross-dataset generalization, we evaluate on 1,512 randomly sampled models from Fusion360 under a standard zero-shot protocol~\cite{khan2024cad-signet}. Fusion360~\cite{willis2021fusion360} is a small CAD reconstruction benchmark with complex and realistic CAD models. Results are shown in Tab.~\ref{tab:fusion360-sota}. Note that some results are borrowed from~\cite{kolodiazhnyi2025cadrille}. Our model consistently outperforms both point-cloud and image-based baselines. For point-cloud reconstruction, \modelname{} achieves the highest IoU (84.3\%), surpassing CAD-Recode~\cite{rukhovich2024cad-recode} (79.1\%) and Cadrille~\cite{kolodiazhnyi2025cadrille} (79.8\%). For image-based reconstruction, we compare with a strong pipeline baseline that converts multi-view images to point clouds using LRM~\cite{hong2024lrm} and then applies CAD-Recode. Even against this optimized hybrid pipeline, \modelname{} improves IoU by +21.4\% (83.9\% vs. 62.5\%) and reduces CD by 3.9× (0.16 vs. 0.62). Compared to Cadrille~\cite{kolodiazhnyi2025cadrille}, our model still achieves better geometry accuracy in both modalities, demonstrating superior generalization despite learning from a more diverse multimodal training objective.

\begin{table}[]
\captionof{table}{Results on Fusion360 test set.
}
\resizebox{\linewidth}{!}{
\setlength{\tabcolsep}{15pt}
\begin{tabular}{l|cc|cc}
\toprule
\multirow{2}{*}{Method} &  \multicolumn{2}{c|}{Point Cloud} & \multicolumn{2}{c}{Multi-view Images} \\
 & CD\textdownarrow & IoU\textuparrow & CD\textdownarrow & IoU\textuparrow  \\
\midrule
DeepCAD~\cite{wu2021deepcad}  & 89.2 & 39.9  & - & -  \\
MultiCAD~\cite{ma2023multicad} & 42.2 & - & - & -  \\
HNC-CAD~\cite{xu2023hnc-cad}   & 36.8 & 63.5 & - & -  \\
TransCAD~\cite{dupont2024transcad}   & 33.4 & 60.2 & - & -  \\
CAD-Diffuser~\cite{ma2024cad-diffuser} & 3.85 & 63.2 & - & -  \\
CAD-SIGNet~\cite{khan2024cad-signet}  & 0.70 & 58.4 & - & -  \\
CAD-Recode~\cite{rukhovich2024cad-recode}  & 0.19 & 79.1  & - & -  \\
LRM~\cite{hong2024lrm}→CAD-Recode~\cite{rukhovich2024cad-recode} & - & - & 0.62 &  62.5  \\
Cadrille~\cite{kolodiazhnyi2025cadrille} &   0.19 & 79.8  &  0.20 & 77.6  \\
\rowcolor{ourscolor}
\textbf{\modelname} &  \textbf{0.18} & \textbf{84.3} &  \textbf{0.16} & \textbf{83.9} \\
\bottomrule
\end{tabular}%
}
\label{tab:fusion360-sota}
\end{table}

\paragraph{Multimodal-Conditioned CAD Generation}
We evaluate the robustness of our model under varying input modalities, as reported in Table~\ref{tab:multimodal-input}. Single-view sketches often capture only partial geometry, leading to incomplete reconstructions. Conditioning on text descriptions mitigates this limitation by providing complementary semantic priors, enabling the model to infer missing geometric regions and generate more complete CAD structures. Fusing sketch and textual inputs consistently outperforms unimodal baselines across both metrics, reducing reconstruction error (CD) and improving surface alignment (IoU). These findings validate the efficacy of our model in leveraging cross-modal signals, particularly when navigating partial or ambiguous observations, resulting in more faithful and geometrically integral CAD reconstructions. By facilitating complementary supervision within a unified executable CAD space, multimodal inputs consistently outperform unimodal baselines. This synergistic fusion not only bolsters model robustness but also aligns with thehybrid-input-driven workflowsprevalent in industrial CAD practice.

\begin{table}[t]
    \captionof{table}{Multimodal-Input-Conditioned CAD Generation.}
    \centering
    \resizebox{0.6\linewidth}{!}{
    \setlength{\tabcolsep}{15pt}
    \begin{tabular}{l|cc}
    \toprule
    Modality & CD\textdownarrow & IoU\textuparrow \\
    \midrule
    Sketch only & 0.38 & 74.2 \\
    Text only & 0.20 & 86.3  \\
    Text + Sketch & \textbf{0.18} & \textbf{88.6} \\
    \bottomrule
\end{tabular}}
\label{tab:multimodal-input}
\end{table}

\section{Conclusion}
We presented UniCAD, a unified benchmark for multi-modal and multi-task CAD understanding and generation. UniCAD standardizes data organization, task definitions, and evaluation across diverse modalities, \ie text, images, sketches, and point clouds, thereby enabling systematic study of text/image/sketch/point-cloud$\rightarrow$CAD, and CAD question-answering tasks within a single framework.
The proposed UniCAD-MLLM achieves state-of-the-art performance and demonstrates strong cross-modal generalization. UniCAD lays a solid foundation for general-purpose CAD intelligence, bridging multi-modal perception and executable design generation.

\paragraph{Limitations and Future Work.}
Despite its contributions, several limitations remain. The current dataset, like existing benchmarks such as DeepCAD and Text2CAD, is constrained by publicly available CAD data and remains largely focused on sketch-and-extrude modeling, with limited coverage of advanced operations such as chamfers, fillets, and freeform surfaces. In addition, the current MLLM backbone is pretrained on natural images rather than CAD-specific geometry, and the point encoder lacks large-scale geometric pretraining, which may limit spatial reasoning for complex structures. Future work will explore CAD-oriented pretraining, geometry-aware fusion, and self-corrective feedback, while extending UniCAD toward industrial datasets and interactive CAD editing scenarios.

\paragraph{Broader Impact.}
UniCAD contributes toward bridging multimodal learning and computer-aided design, promoting interpretable and reproducible 3D model generation. Its deterministic code-based output may benefit manufacturing, reverse engineering, and digital twin applications by improving design automation and knowledge transfer. However, as with all generative models, risks include potential misuse for proprietary design reproduction or biased dataset artifacts. We encourage responsible deployment of UniCAD, accompanied by dataset transparency, open evaluation, and adherence to ethical design and intellectual property standards.

{
    \small
    \bibliographystyle{ieeenat_fullname}
    \bibliography{main}
}


\newpage
\appendix

\section{Details on UniCAD Dataset Construction}

\subsection{Sketch Generation}
We input the multi-view rendered images of CAD models into the image editing model from Qwen-Image-Edit~\cite{wu2025qwen}. The following prompt was used: 
\begin{quote}
\textit{``This is a multi-view rendered image of a CAD part. Please convert the style to a sketch style while preserving the original image content. Only modify the visual style and keep the geometry and structure unchanged.'' }
\end{quote}
This process enables us to obtain sketch-style images that maintain the original geometric information of the CAD models while providing a hand-drawn visual appearance.

\subsection{Text Description Generation}
We adopt a two-stage strategy for generating textual descriptions of CAD models.

\textbf{Stage 1: Abstract Geometric Feature Synthesis.}
In the initial stage, we leverage GPT-4o to generate abstract, object-level textual descriptions that encapsulate the visual and structural essence of CAD models. This process is engineered to produce meticulous geometric reports, capturing intricate features such as hollow frustums with uniform wall thickness, hemispherical shells with internal vertical ribs, or T-shaped structures with specific hook-like protrusions.

To achieve this, each CAD model is first rendered into multi-view images. These images are then fed into GPT-4o using a specialized Structured Engineering Prompt:

\begin{quote}
\textit{``As a Senior CAD Engineer, provide a rigorous geometric decomposition report to a junior designer, enabling them to precisely reconstruct the model without visual assistance. Based on the multi-view rendered images, integrate the primitive geometric shapes, profile and cross-sectional features, 3D and local structural characteristics (such as draft angles, fillets, and ribs), topological structure, and spatial relationships between components into a standardized modeling logic. Ensure the description covers the complete technical workflow, ranging from initial sketch constraints to advanced feature modeling, including Boolean operations, patterns, and shelling.''}
\end{quote}

By following this prompt, GPT-4o generates detailed natural language descriptions that prioritize geometric and constructive integrity. This approach ensures the output focuses on the model's fundamental topology while intentionally filtering out irrelevant visual attributes such as color or material texture.

\textbf{Stage 2: High-Fidelity Textual Description Generation.}
In the second stage, we generate detailed textual instructions that describe the CAD construction process. Specifically, we integrate the CadQuery~\cite{cadquery} source code, multi-view rendered images, and the geometric descriptions generated in the first stage as multimodal inputs to GPT-4o. This integration aims to produce high-fidelity, expert-level CAD modeling specifications, rather than merely providing simple visual summaries of the models. By incorporating precise geometric topology, absolute and relative dimensional parameters, and rigorous sketch constraints, the generated text clearly conveys the underlying parametric construction logic of the model, enabling users to better understand the modeling process and supporting high-precision reconstruction of the CAD model.

To achieve this, we have developed a systematic conversion framework guided by a Structured Expert Prompt:

\begin{quote}
\textit{``You are a Senior Lead CAD Engineer. Your task is to translate parametric CAD designs into highly precise, natural language modeling instructions to guide high-fidelity reconstruction in any professional CAD software. Based on the provided CadQuery code, rendered images, and initial geometric descriptions, you must:
(1) Extract all critical numerical parameters (e.g., radii, lengths, extrusion depths) and define their relative proportions;
(2) Explicitly identify reference workplanes (XY, YZ, XZ) and the origin;
(3) Precisely specify target entities such as 'the topmost face' or 'edges parallel to the X-axis';
(4) Define sketch constraints including concentricity, tangency, and axial symmetry.
Integrate dimensions directly into the geometric description while strictly omitting any non-functional visual attributes such as color, material, or lighting.
The resulting output is organized into a standardized, five-part architectural hierarchy: (1) Global Parameters and Constants: Definition of key variables and their proportional relationships.
(2) Reference Planes and Primary Geometry: Sketch constraints and foundational 3D modeling operations.
(3) Secondary Features and Boolean Operations: Precise positioning and sizing of additive or subtractive structures.
(4) Transformation and Pattern Logic: Detailed descriptions of linear/circular arrays or mirroring operations.
(5) Topological Refinement: Identification of target edges and parameters for fillets and chamfers.''}
\end{quote}

Through this structured workflow, the system produces highly detailed, parameter-driven CAD modeling descriptions, effectively bridging the gap between CAD textual representations and engineering design intent. For example, for the three CAD models in Stage 1, the following descriptions are generated:
\begin{itemize}
\item Component 1: Tapered Hollow Cylinder Array specifies global parameters including outer radius $R_{out} = 0.4$, inner radius $R_{in} = 0.3$ (wall thickness $w = 0.1$), extrusion height $H = 0.8$, negative draft angle $\alpha = -5^\circ$, and linear array pitch $S = 0.9$. On the XY workplane at the origin, concentric circle sketches define the cylinder cross-section, extruded along the +Z direction with the taper applied. A secondary base plate of thickness $0.05$ is integrated via Boolean union at the bottom face, followed by a $2 \times 2$ linear array along the X and Y axes to complete the configuration.
\item Component 2: Ribbed Hemispherical Shell Array has a sphere radius $R_s = 0.5$, shell thickness $t = 0.04$, and internal ribs with width $W_r = 0.06$ and depth $D_r = 0.12$. A semicircular sketch on the XZ plane is revolved and shelled to form a hemispherical shell. Three rectangular ribs are added on the interior surface with $120^\circ$ radial symmetry and tangency constraints, and then arranged in a circular array of four units around the global origin, with $90^\circ$ intervals. 
\item Component 3: Orthogonal T-Hook Component Array defines a post height $H_p = 0.6$, crossbar width $W_b = 0.4$, hook extension $L_h = 0.15$, and extrusion thickness $T = 0.08$. A T-shaped profile is sketched on the YZ plane with perpendicularity and axial symmetry constraints, extruded symmetrically along the X-axis. An L-shaped hook with a fillet radius of $0.03$ is extruded from the bottom of the vertical post, and four instances are placed via rotational symmetry around the Z-axis, with hooks facing inward toward the center. All dimensions, topological references, and sketch constraints are explicitly defined, while non-functional visual attributes such as color or material are omitted.
\end{itemize}

\subsection{QA Pairs Generation.}
We input the multi-view rendered images and textual descriptions into GPT-4o, and by using a structured prompt, we guide the model to generate a series of four-option multiple-choice questions (MCQs) to assess a comprehensive understanding of CAD geometry. The prompt is formulated as follows:

\begin{quote}
\textit{``Given the multi-view rendered images and the structural description of the CAD model, generate a series of geometric assessment questions. Each question must include four answer options and a correct answer, covering the following domains: (1) Fundamental Geometry—primitive types, surface and curve characteristics, and features such as holes or bosses; (2) Topology—entity relationships, face-edge connectivity, and structural hierarchy; (3) Constraints—sketch-level constraints and 3D spatial alignment; and (4) Patterns—transformation logic, including linear and circular arrays as well as mirror symmetry. The output format should follow the template: Q: [question], A: [correct answer].''}
\end{quote}

By leveraging this integrated multimodal framework, large-scale, high-fidelity question-answering datasets for CAD models can be generated, providing a solid foundation for the training and benchmarking of specialized LLMs in the CAD model domain.

\section{Comparison and Rationale for Selecting CadQuery}

\subsection{Comparative Analysis}
We conducted a comparative analysis of four parametric CAD modeling tools. The evaluation prioritizes practical attributes critical to automated workflows: software dependencies, scripting language accessibility and modeling complexity (Table \ref{tab:cad-tool-comparison}).

\begin{table}[htb]  
    \centering
    \caption{Comparison of Open-Source CAD Modeling Tools}
    \label{tab:cad-tool-comparison}
    \begin{tabular}{@{}lccc@{}}  
        \toprule
        \textbf{Tool} 
        & \textbf{\makecell{Requires\\Software}} 
        & \textbf{\makecell{Scripting\\Language}} 
        & \textbf{\makecell{Modeling\\Complexity}} \\
        \midrule
        FreeCAD      & Yes & Python   & Low    \\
        OpenSCAD     & Yes & OpenSCAD DSL & Medium \\
        PythonOCC    & No  & Python   & High   \\
        CadQuery     & No  & Python   & Low    \\
        \bottomrule
    \end{tabular}
\end{table}

\begin{figure*}
  \centering
  \includegraphics[width=0.9\textwidth]{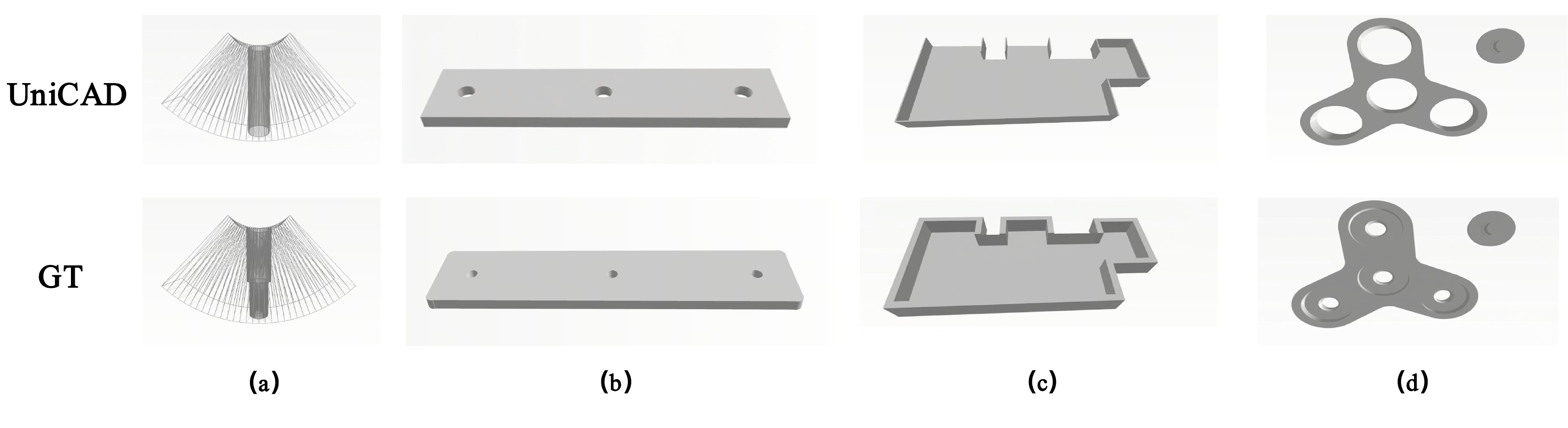}
  \caption{Failure cases of CAD reconstruction on DeepCAD dataset.}
  \label{fig:Failure cases}
\end{figure*}

\begin{figure*}
  \centering
  \includegraphics[width=0.7\textwidth]{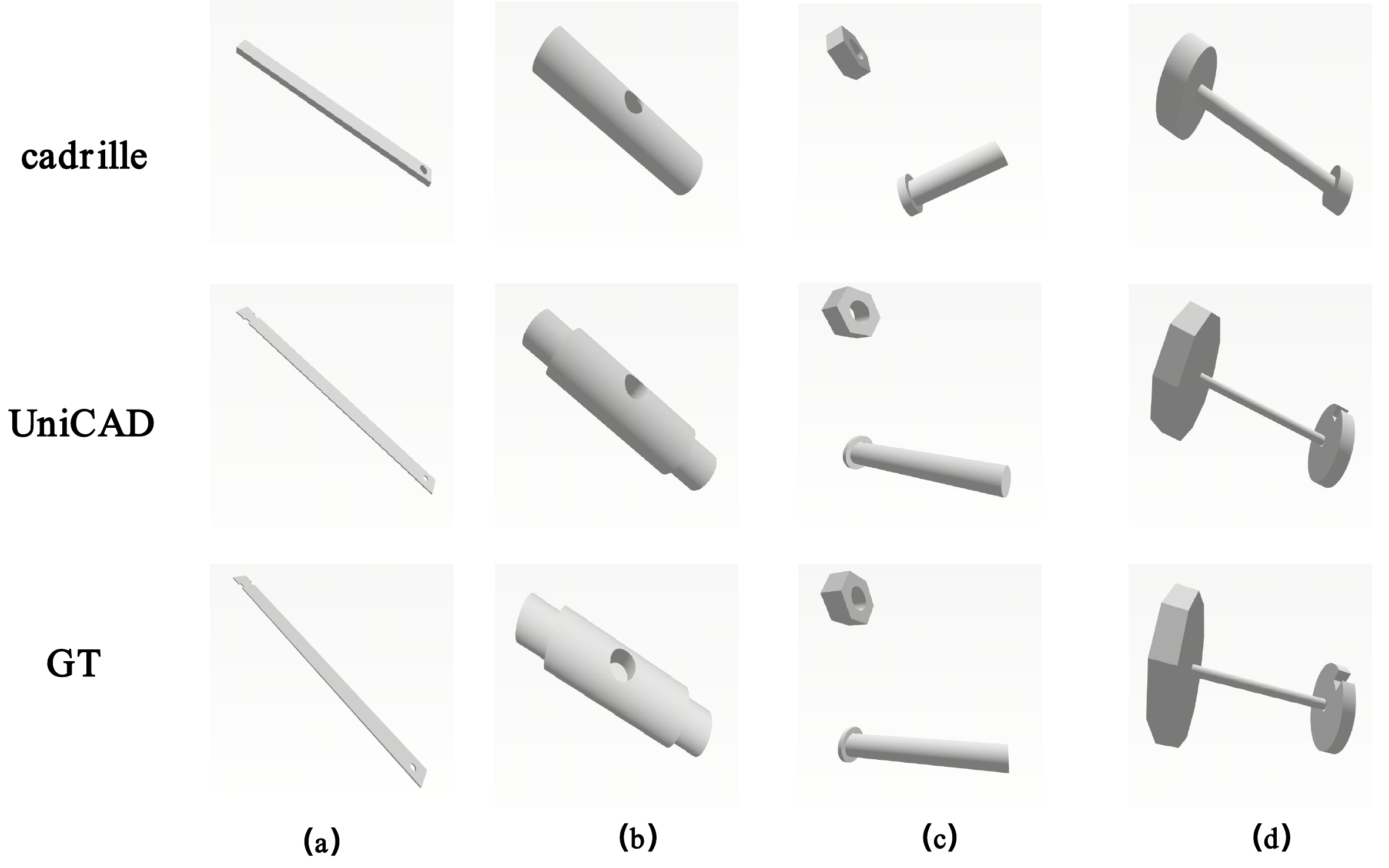}
  \caption{Comparison of CAD Reconstruction Results between UniCAD and cadrille.}
  \label{fig:Comparison}
\end{figure*}

Among these tools, CadQuery is uniquely suited to our automated parametric modeling pipeline. It eliminates mandatory external software dependencies (unlike FreeCAD/OpenSCAD) while retaining Python as its native scripting language—avoiding the proprietary DSL limitations of OpenSCAD and the steep learning curve of PythonOCC’s low-level geometric kernel interface. CadQuery’s key advantage lies in its balance of simplicity and automation readiness: it offers a high-level, intuitive API (matching FreeCAD’s low modeling complexity) without being tethered to a GUI (FreeCAD’s critical limitation for batch processing) or requiring expertise in low-level geometric operations (PythonOCC’s primary barrier). For our use case, programmatic, scalable shape synthesis, CadQuery’s script-native design (no GUI locks, no external software requirements) and low modeling complexity make it the optimal choice.

\subsection{Modeling Examples}
To illustrate the practical differences in syntax and workflow for a basic 10×10×10 cube, we provide minimal, executable code snippets for each tool:

\paragraph{FreeCAD (GUI-Dependent Python API)}
FreeCAD requires a GUI context and explicit document management, making batch automation cumbersome:
\begin{lstlisting}[language=Python, 
                   basicstyle=\small\ttfamily,
                   keywordstyle=\color{blue},
                   commentstyle=\color{green!60!black},
                   frame=single,
                   caption={FreeCAD: 10×10×10 Cube Generation},
                   label={lst:freecad-cube}]
import FreeCAD, Part
# Create a new GUI document
doc = FreeCAD.newDocument("Example")
# Generate cube geometry
box = Part.makeBox(10, 10, 10)
# Render in GUI
Part.show(box)
# Refresh GUI to reflect changes
doc.recompute()  
\end{lstlisting}

\paragraph{OpenSCAD (Proprietary DSL)}
OpenSCAD uses a custom domain-specific language (DSL) with no native batch automation support:
\begin{lstlisting}[language=Scala, % Fallback for OpenSCAD syntax
                   basicstyle=\small\ttfamily,
                   keywordstyle=\color{blue},
                   commentstyle=\color{green!60!black},
                   frame=single,
                   caption={OpenSCAD: 10×10×10 Cube Generation},
                   label={lst:openscad-cube}]
// 10x10x10 cube (OpenSCAD DSL)
cube([10, 10, 10]);
\end{lstlisting}

\paragraph{PythonOCC (Low-Level Python API)}
PythonOCC requires direct interaction with the OpenCASCADE kernel, introducing unnecessary complexity for basic modeling:

\begin{lstlisting}[language=Python,  
                   basicstyle=\small\ttfamily,
                   keywordstyle=\color{blue},
                   commentstyle=\color{green!60!black},
                   frame=single,
                   caption={PythonOCC: 10×10×10 Cube Generation},
                   label={lst:pythonocc-cube}]
from OCC.Core.BRepPrimAPI import BRepPrimAPI_MakeBox 
from OCC.Display.SimpleGui import init_display

# Create cube geometry
box = BRepPrimAPI_MakeBox(10, 10, 10).Shape()
# Initialize graphical display
display, start_display, _, _ = init_display()
# Render cube and refresh view
display.DisplayShape(box, update=True)
# Start display loop
start_display()
\end{lstlisting}

\paragraph{CadQuery (Script-Native Python API)}
CadQuery enables concise, automation-ready modeling with no external dependencies or GUI locks:
\begin{lstlisting}[language=Python,  
                   basicstyle=\small\ttfamily,
                   keywordstyle=\color{blue},
                   commentstyle=\color{green!60!black},
                   frame=single,
                   caption={CadQuery: 10×10×10 Cube Generation},
                   label={lst:cadquery-cube}]
import cadquery as cq
# Create cube geometry
box = cq.Workplane("XY").box(10, 10, 10)
\end{lstlisting}

\section{Subjective A/B Test for Sketch Authenticity}

To verify the authenticity and task practicability of our generated sketches, we conducted a blind A/B test with 8 participants (3 professional CAD designers and 5 Phd students). Each participant was presented with 100 blind test images, and each image was randomly selected from one of the three sources: a real hand-drawn sketch from Free2CAD, a synthetic sketch generated by UniCAD, and a line drawing produced by Canny edge detection. All images were standardized in format with their source information concealed.
For the rating task, participants were instructed to score each image on a 1–5 Likert scale (1 = Poorest, 5 = Best) in terms of authenticity and correctness. The results show that our synthetic sketches achieved an average authenticity score of 4.3/5, which is close to the 4.5/5 score of real hand-drawn sketches and significantly higher than the 1.8/5 score of edge-detected line drawings.

\section{Qualitative Comparisons}

\subsection{Failure Cases}
Figure~\ref{fig:Failure cases} illustrates several failure cases. For example, in Figure~\ref{fig:Failure cases}(a), the ground truth is a sector-shaped part with a cylindrical hole that is wider at the top and narrower at the bottom, whereas our reconstruction produces a uniformly sized hole. In Figure~\ref{fig:Failure cases}(b), the reconstructed part lacks the chamfer present in the ground truth. In Figure~\ref{fig:Failure cases}(c), the reconstructed part is thinner and misses the closed modeling feature. In Figure~\ref{fig:Failure cases}(d), the hole in the reconstructed part is uniform in diameter, differing from the tapered hole in the ground truth. Nevertheless, our method UniCAD generally produces geometrically plausible shapes, though it still struggles to accurately recover fine-grained geometric details for objects with complex surface structures. Overall, the reconstructed parts largely preserve the correct geometric shape, deviating from the ground truth only in fine-grained details.

\subsection{Visualization Results}
We compare UniCAD with the previous state-of-the-art method, cadrille~\cite{kolodiazhnyi2025cadrille}, on the CAD reconstruction task, presenting the visualization results in Figure~\ref{fig:Comparison}. UniCAD successfully reconstructs all the models shown in Figure~\ref{fig:Comparison}, while Cadrille fails to accurately recover several of them. Specifically, in Figure~\ref{fig:Comparison}(a), the target model is reconstructed as a cuboid with excessive thickness; in Figure~\ref{fig:Comparison}(b), it erroneously reconstructs the object as a cylinder; in Figure~\ref{fig:Comparison}(c), it fails to recover the correct dimensions of the screw component; and in Figure~\ref{fig:Comparison}(d), both the upper and lower parts of the model are reconstructed as cylindrical structures. The results demonstrate that UniCAD achieves superior CAD reconstruction performance compared with cadrille, particularly in preserving the geometric structures of the target models. More importantly, UniCAD demonstrates stronger sensitivity to dimensional parameters, allowing it to more accurately preserve both the scale and size information of reconstructed CAD models.

\end{document}